\documentclass[conference]{IEEEtran}
\usepackage{subfigure}
\usepackage{amsmath}
\usepackage{graphicx}
\usepackage{epsfig}
\usepackage{float}
\usepackage{cite}

\ifCLASSINFOpdf
\else
\fi
\begin{document}
%
\title{Deeply-Supervised CNN for Prostate Segmentation}

\author{
\IEEEauthorblockN{Qikui Zhu}
\IEEEauthorblockA{School of Computer\\
 Wuhan University\\
 WuHan, China,430079\\
 Email: QikuiZhu@163.com}
\and
\IEEEauthorblockN{Bo Du}
\IEEEauthorblockA{School of Computer\\and State Key
Lab of Information\\ Engineering on Survey,Mapping and Remote Sensing\\
 Wuhan University\\
WuHan, China,430079\\
Email: gunspace@163.com}
\and
\IEEEauthorblockN{\hspace{ 2cm }Baris Turkbey and Peter L . Choyke}
\IEEEauthorblockA{\hspace{ 2cm }National Cancer Institute\\\hspace{ 2cm }National Institutes of Health
\\\hspace{ 2cm }9000 Rockville Pike, Bethesda, MD\\Email: Ismail.Turkbey@nih.gov, pchoyke@mail.nih.gov}
\and
\IEEEauthorblockN{\hspace{ 2cm } Pingkun Yan}
\IEEEauthorblockA{\hspace{ 1.5cm } Philips Research, 10 Center Driv, Bethesda
\\\hspace{ 2cm }MD,20892,USA
\\\hspace{ 2cm }Email: pingkun.yan@philips.com}
}


%


\maketitle

\begin{abstract}
Prostate segmentation from Magnetic Resonance (MR) images plays an important role in image guided intervention.
However, the lack of clear boundary specifically at the apex and base, and huge variation of shape and texture
between the images from different patients make the task very challenging. To overcome these problems, in this paper, we propose a
deeply supervised convolutional neural network (CNN) utilizing the convolutional information to accurately segment the prostate from MR images.
The proposed model can effectively detect the prostate region with additional deeply supervised layers compared with other approaches.
Since some information will be abandoned after convolution, it is necessary to pass the features extracted from early stages to later stages.
The experimental results show that significant segmentation accuracy improvement has been achieved by our proposed method compared to other reported approaches.
\end{abstract}


%
\IEEEpeerreviewmaketitle

\section{Introduction}
 Over the past few years, the capabilities of Convolutional Neural Networks (CNNs) have achieved state-of-the-art performances in many fields, such as computer vision\cite{LeaningDe} and medical
  image analysis\cite{Baxt1992Use}. This success lies in the following aspects\cite{Szegedy2014Going}: (1) more powerful graphics processing units (GPUs) have been developed; (2) a huge amount of available data, for example, about 1.2 million annotated images were provided in ImageNet Large Scale Visual Recognition Challenge (ILSVRC)\footnote{http://www.image-net.org/challenges/LSVRC/2014/}. (3) Many networks have been designed for specific tasks, such as classification\cite{Krizhevsky2012ImageNet,Simonyan2014Deep,Dan2012Multi}, segmentation\cite{FULL,Zhang2015Deep} and object detection\cite{Simonyan2014Very,He2014Spatial}.
 The core ability of CNNs is to learn hierarchical representation of the data, so adjusting networks' structure to improve the ability of hierarchical representation is a main objective in CNN based applications.\vspace{2mm}

 Classification is the most common application in CNNs, such as GoogLeNet\cite{Szegedy2014Going}, VGG-Net\cite{Simonyan2014Very}, where the output is a class
 label for the image. However, in many visual tasks, especially in medical image analysis\cite{Yan2011Adaptively,Dan2012Deep}, specific needs have to be met.
 For example, in medical image segmentation, the label is supposed to be assigned to each pixel and the result should have high precision. With impressive improvements by deep learning, more and more researchers apply these approaches  in different medical image applications, such as image segmentation\cite{Zhang2015Deep,Kleesiek2016Deep}, image fusion\cite{Suk2014Hierarchical}, image registration\cite{Wu2016Scalable} and computer-aided diagnosis\cite{Cire2013Mitosis,Drozdzal2016The}.\vspace{1mm}
\begin{figure}
\centering
\subfigure[]{
\includegraphics[width=2.4cm,height=2.0cm]{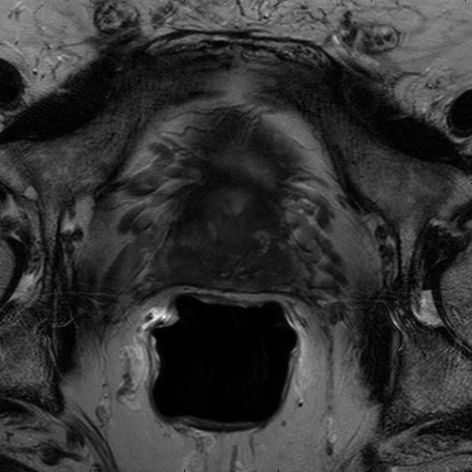}
}
 \subfigure[]{
\includegraphics[width=2.4cm,height=2.0cm]{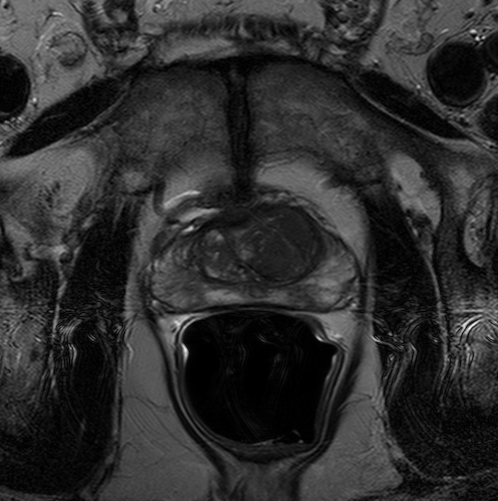}
}
 \subfigure[]{
\includegraphics[width=2.4cm,height=2.0cm]{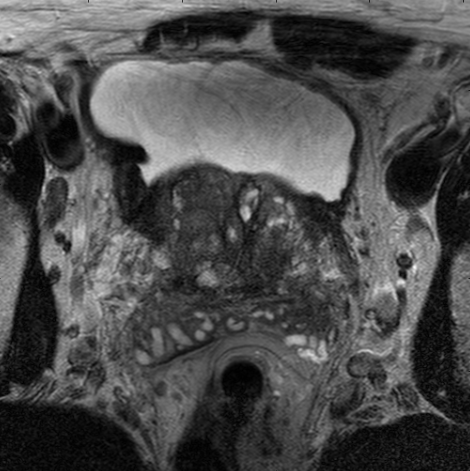}
}
\caption{Challenges in segmenting prostate in MR images. (a) Imaging artifacts inside the bladder. (b) Weak boundary. (c) Complex intensity distribution.}
\end{figure}
Automatic segmentation is one of the pillars of medical image analysis. The key to segment medical images successfully is highly dependent on edge detection in
 a given context. So obtaining the edge's features and then searching for its position is the main work in many methods. Over the past decades, constructing
 effective feature engineering\cite{Lefei} has been a mainstream topic in medical image segmentation. For instances, Shen et al.\cite{Shen2001HAMMER} proposed the geometric
  moment invariant based features for feature-guided non-rigid image registration, and Liao et al.\cite{Liao2013Representation} proposed a representation learning
  method for automatic feature extraction in medical image analysis and a stacked dependent subspace analysis (ISA) deep learning framework is proposed to automatically
  learn the most informative features from the input image. Besides, shape-based models are widely used for image segmentation. Yan et al. \cite{Yan2010Discrete} proposed a
  method of utilizing a prior shape estimated from partial contours for segmenting the prostate. Toth et al. \cite{Toth2012Multifeature} constructed  an AAA model by utilizing
  the intensity and gradient information, and then utilizing level-set method to segment prostate MRI. All of these methods segment the medical images by utilizing the specific
  features information. However, as we have described above, deep learning possesses good ability of learning hierarchical feature representations from data and
  has achieved record-breaking performance in a variety of applications. In terms of medical image analysis, which highly depends on edge detection. The main work in many methods is to find edges of the structure of interest.
  And deep learning can learn edge feature effectively. For
instance, Xie and Tu\cite{Xie2015Holistically} proposed a convolutional neural network based edge detection system for edge and object boundary detection. So we believe that deep learning can achieve great improvement in medical image analysis just like in
  computer vision. Many researchers have utilized deep learning in medical image analysis. For instances, Zhang et al. \cite{Zhang2015Deep} proposed to use deep convolutional neural networks (CNNs) for segmenting
 isointense stage brain tissues using multi-modality MR images. Cheng et al.\cite{Cheng} proposed
 a supervised machine learning model which utilized atlas based Active Appearance Model and a deep
 learning model to segment the prostate on MR images. Chen et al. \cite{Chen2016DCAN} proposed a deep
 contour-aware network that integrates multi-level contextual features to segment glands. All of
  these methods have utilized the advance of deep learning and obtained outstanding performances. \vspace{2mm}
\begin{figure*}
\centering

\includegraphics[width=10cm,height=6cm]{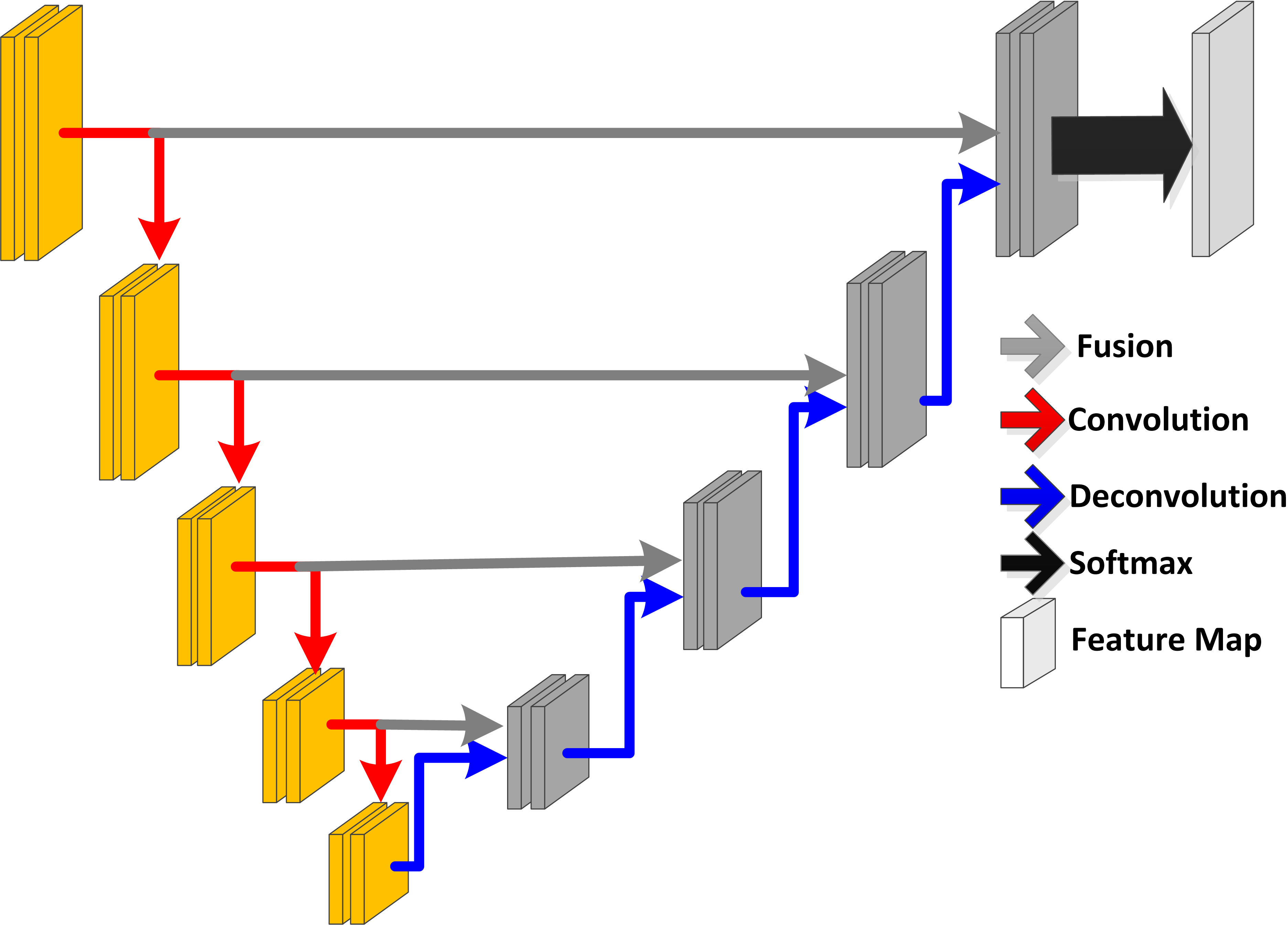}
\caption{ U-net architecture. Each box corresponds to a multi-channel features maps. The number of features maps increases stage by stage on the left part. on the contrary, the number of features maps decreases stage by stage on the right part. The horizontal arrow denotes transfer residual information form early stage to later stage.}
\end{figure*}

Many networks apply patch-to-pixel or patch-to-patch strategy to train and predict. However, this strategy always results in significantly downgraded training and prediction efficiency.
The Fully Convolutional Neural networks (FCNs)\cite{Long2015Fully} provide a way to train the network image-to-image, which allows us to train on a larger amount of samples simultaneously. However, we cannot directly apply FCN in prostate segmentation.
Due to the prostate always lacks of clear boundary specifically at the apex and base, and the shape and texture are huge variation between different patients. These phenomenons make the prostate segmentation become a challenge.
Inspired by these methods and the superiority of deep learning, we propose a network which can forward the
features extracted from early stages to later stages to avoid information lost. And to keep the features produced at hidden layers semantically
meaningful, we put additional deeply supervised layers\cite{Lee2014Deeply} at each stage. We name the proposed network as Deeply-Supervised CNN which, trained end-to-end, can segment the  prostate prostate on MR images accurately
and fast. Our network has three stages, the first stage consists
of a compression path which extracts features from the data and reduce the resolution by an appropriate
 stride. The second stage of the network consists of an expansive path which upsamples the feature map
  and halve the number of feature channels until its original size is reached. In order to help the network learn more
  precise residual information, the third stage is constructed by deep supervised layers which supervise the process of training.

\begin{figure*}
\centering

\includegraphics[width=16cm,height=10cm]{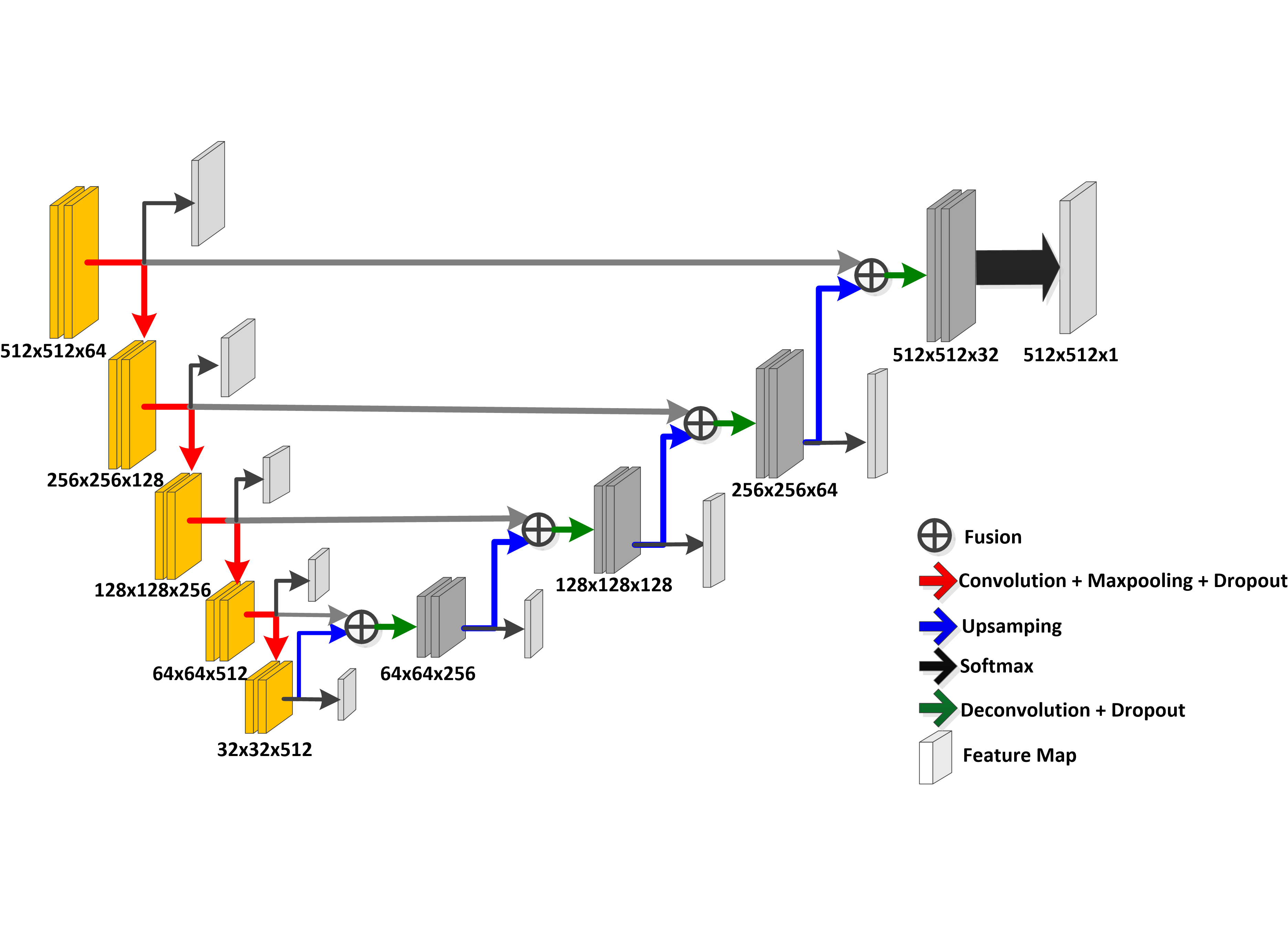}
\caption{ Deeply-supervised CNN architecture.Each box corresponds to a multi-channel feature map.The number of feature map increase stage by stage on the left part. on the contrary, the number of feature map decrease stage by stage on the right part. The gray arrow denotes transfer residual information form early stage to later stage. At each stage, we add s supervised layer.}
\end{figure*}
\section{Methods}
\subsection{U-Net}
The architecture of U-Net \cite{Ronneberger2015U} is illustrated in Fig 2. This contains two parts, the left part of the network is divided into four stages.
 Each stage consists of two convolutional layers and deals with  different resolution feature maps. The convolution performed in left part uses 3x3 kernels,
 and each is followed by a rectified liner unit (ReLU)\cite{Gu2015Recent}. And at the end of each stage, a 2x2 max pooling operation with stride of 2 is attached for down sampling. The number of feature channels is doubled after each stage.
 The right part of the network is also divided into four stages. The architecture of the right part is similar to
the left. Each stage of right part includes two kinds of operations. The first is
upsampling which makes the size of feature map increase gradually until it reaches
the size of the original input image. The second operation is to halve the number of feature channels, so
that the number of convolution kernels will be halved after each stage. Since some image information will be lost
after every convolution, it is necessary to put forward the features extracted from early stages
of the left part to the right part. In order to achieve this function, the authors makes the left parts connected
with the right parts.  In this way, the network can gain some details that otherwise have been lost during convolution.
And this operation will improve the quality of the final contour prediction. Besides these connections
will speed up the convergence of the network.

\subsection{Deeply-Supervised CNN}
\subsubsection{Network Architecture}
In this section, we describe the details of the proposed network's architecture.
As VGG-Net\cite{Simonyan2014Very}  has demonstrated that the representation depth is beneficial for classification accuracy. In order to obtain higher accuracy, it would be beneficial to utilize deeper network to segment
 prostate images. However, deeper network also brings two bottlenecks. First, deeper network typically means
 larger number of parameters, which make the network more prone to overfitting, especially for the application
 of medical images, because the number of labeled examples in the training set is always limited. The other bottleneck
 of deeper network is the dramatically increased use of computational resources. To solve this problem, we propose to
 use 1x1 convolutional layers in the convolutional process. The 1x1 convolution has two major advantages. On the one hand,
 they can reduce the dimensions and the number of parameters, remove computational bottlenecks to some extent; on the other
 hand, they can increase the depth of the network and improve the ability of character representation. In our network,
 we have applied 1x1 convolution in many stages to improve the accurate of segmentation. \vspace{2mm}

As shown by GoogLeNet\cite{Szegedy2014Going} that smaller convolutional kernels are more efficient in 2D network and smaller
convolutional kernels can get the same effects as that large kernel gets. It can be proved that the effective receptive field
size of stacked small kernels is equivalent to that of one large kernel. In addition, the smaller convolutional kernels can
reduce the number of parameters and remove computational bottleneck simultaneously\cite{Szegedy2014Going}. So in our network, the convolutional
 kernels size are all set as 3x3. Additionally, pooling operations have significance on improving the state-of-the-art convolutional networks and
  are useful to combat overfitting to some extent, so we have added pooling operations at the end of each stage.\vspace{1mm}

As we have described above, the operation of convolution always results in image information lost. By forwarding the features extracted from
early stages to later stages can provide those losing information for later stages. At last, improving the quality of final
prediction. However, this is still leaves some space to be improved. When we forward features from early stages to later stages,
due to lack of deep supervision, the features produced at hidden layers are less semantically meaningful. More importantly, they
 will significantly influence training and prediction efficiency. However, as Deeply-Supervised Nets\cite{Lee2014Deeply} has demonstrated that deeply supervised
 layers can improves the directness and transparency of the hidden layer learning process. To overcome the described problems, we put eight additional deep supervised
 layers in the network. During training, all of these supervised layers supervise the process of training. Some times, since the depth of the
 network is large, the ability to propagate gradients back through all the layers in an effective manner is a concern. The additional supervised
  layers can solve the problem well by preserving gradients from early stage. \vspace{2mm}

In summary, the proposed network contains three parts as shown in Fig 3. The first five stages
consist of a compression path which extract features from the data and reduce the resolution by an appropriate stride. From top to
the fourth stage, the number of feature channels is doubled at each stage. In the first stage, the number of feature channels is 64, for example,
after four stages the number of feature channels increased to 512. In each stage, we perform two 3x3 convolutions, one 1x1 convolution and
one 2x2 max pooling operation with stride 2 is attached for down sampling. On the contrary,
the later four stages consist of an expansive path which upsample the features maps and halve the number of feature channels until its original
 size is reached. These stages have same operates like the stage within compression path except the max pooling operation. On the part of supervised
 layers, each supervised layer consists of a upsampling layer and a deconvolution layer. The upsampling layer upsamples the features map and then via
 deconvolution layer to obtain the segment result. During training, these supervised layers control the
 process of training according to the difference between segmentation result and ground truth .\vspace{2mm}

The proposed network has some superiority over the original U-net. For instance, the 1x1 convolutions make the network become deeper as well as be not
get into computational difficulties. And the operator of max pooling has significance in improving the state-of-the-art convolutional networks and overcoming
the overfitting. Besides, the additional deep supervised layers make the residual information meaningful and improve the convergence time of the model.
\subsubsection{Formulation}
We denote our input training dataset by $S = \{ ({X_n},{Y_n}),n = 1,...,N\}$, where ${X_n} = \{ x_j^n,j = 1,...,\left| {{{\rm{X}}_n}} \right|\}$ denotes the raw input image and ${{\rm{Y}}_n} = \{ {\rm{y}}_j^n,j = 1,...,\left| {{{\rm{X}}_n}} \right|\}$ denotes the corresponding ground truth binary edge map for image ${X_n}$. For simplicity, we denote all network layers' parameters as ${\rm{W}}$. In deep supervised layers, the corresponding weights are denotes as $
{\rm{w = \{ }}{{\rm{w}}^1}{\rm{,}}...{\rm{,}}{{\rm{w}}^m}{\rm{\} }}$,where ${\rm{m}}$ denotes the number of deep supervised layers (in our method, ${\rm{m}}$= 8). We consider the objective function
\begin{equation}
{{\rm{L}}_{supervised}}{\rm{(W,w) = }}\sum\limits_{i = 1}^m {{\alpha _i}l_{supervised}^i(W,{w^i}) }
\end{equation}

\noindent where ${l_{supervised}}$ denotes the image-level loss function for deep supervised layers' outputs. However, for the prostate images, the anatomy of interest occupies only a very small region of the scan. This always causes the network ignores the segmentation parts and the output of network are quite biased towards background, the learning process get trapped in local minima and can not obtain accurate results finally. To avoid this problem, in this paper, we utilized the dice coefficient as the objective function, which ranges between 0 and 1. The dice coefficient (DSC)\cite{Duan2010A} between two images can be written as
\begin{equation}
DSC({S_a},{S_m}) = \frac{{2\left| {{S_a} \cap {S_m}} \right|}}{{\left| {{S_a}} \right| + \left| {{S_m}} \right|}}
\end{equation}
\noindent where ${S_a}$ denotes the shape of automatic segmentation and ${S_m}$ denotes the shape of manual segmentation.\vspace{2mm}

In our work, the ground truth and result of segmentation are binary image, so the dice coefficient DSC between two binary images can be written as
\begin{equation}
{\rm{DSC}} = \frac{{2\sum\nolimits_i^N {{p_i}{q_i}} }}{{\sum\nolimits_i^N {p_i^2}  + \sum\nolimits_i^N {q_i^2} }}
\end{equation}
\noindent where $N$ denotes the total number of pixels in the image, ${p_i}$ and ${q_i}$ denotes the pixel of ground truth and segmentation respectively.\vspace{2mm}

Applying this formulation in our methods, we do not need to balance the number of samples between foreground and background pixels. Except for the supervised layers, we should also consider the final output. Putting all of loss together, we should minimize the following objective function via standard stochastic gradient descent
\begin{equation}
(W,w) = argmin({L_{supervised}}{\rm{(W,w)}} + L(W,w))
\end{equation}
\noindent where $L(W,w)$ denotes loss function of the final output.

\section{Training}
\subsection{Dataset}
In this work, all of the data is acquired from 81 patients. These images are acquired by a Philips 3T MRI scanner with endorectal coil.
The in plane resolution is 0.3mm x 0.3mm and inter-plane distance is 3mm. Each volume consists of 26 slices and each slice has 512x512 pixels.
When training the network, we remove some slices which do not contain the prostate. So,the total number of images is 1324.\vspace{2mm}

The training and testing samples are randomly selected from the dataset. We selected 4 patients (totally 64 images) for testing and the
rest of patients are utilized for training.\vspace{2mm}

Since we have very limited number of images, it always results in the model suffering from overfitting. To increase the robustness and reduce
 overfitting, we employed the strategy of data augmentation to enlarge the training dataset. The augmentation transformations include translation,
 rotation and zoom.
\begin{figure}
\centering
\subfigure[]{
\includegraphics[width=3cm,height=3cm]{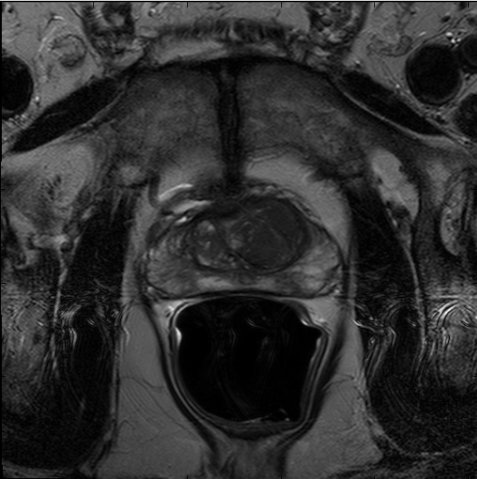}
\includegraphics[width=3cm,height=3cm]{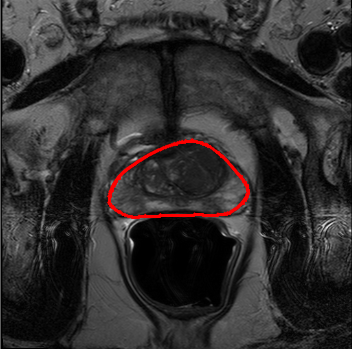}
}

 \subfigure[]{
\includegraphics[width=3cm,height=3cm]{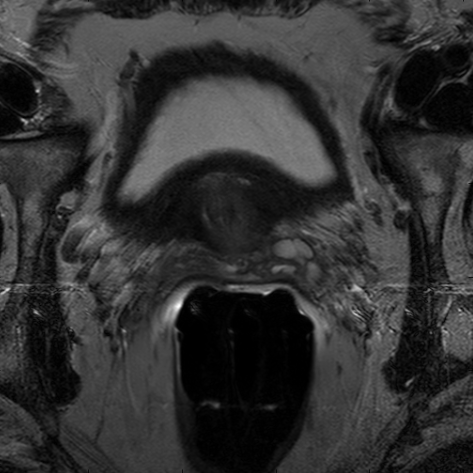}
\includegraphics[width=3cm,height=3cm]{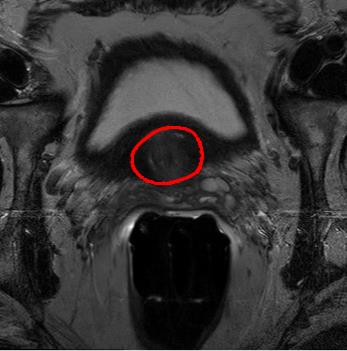}
}
\caption{Input image and the relative manual ground truth annotation. }
\end{figure}
\subsection{Implementation Details}
Our network is trained end-to-end on a dataset of prostate scans in MRI. All of the training images and ground truth have a fixed size of 512x512 and the
framework is implemented under the open-source deep learning library Keras\cite{2016arXiv160502688full}. In the training phase, the learning rate is
 initially set as 0.001 and decreased by a weight decay. The momentum is 0.9, and due to the limitation of the memory, we choose 1 as the batch size.
 Experiments are carried out on GTX1080 GPU with 8 GB of video memory and the CUDA edition is 8.0.\vspace{2mm}

\section{Result and Discussion}
We trained our network on 77 patients. The input images and the manual ground truth annotation are shown in Fig.4.
As we have described above, these images were acquired from different patients, and these images include the
clinical variability. To evaluate our method, we randomly selected 4 patients with  64 images before
training. These images do not take part in training and the prostate has been manually pre-delineated by a radiologist, which were used
as the ground truth to evaluate the performance of automatic segmentation. We also selected dice coefficient
as the evaluation method\cite{Duan2010A}. And to validate our method against U-Net and fully convolutional networks (FCNs), we used the same dataset to
train and test the U-Net and FCNs. \vspace{2mm}
\subsubsection{Qualitative Comparison}
To intuitively compare the proposed method with U-Net and FCN, the segmentation results of some representative and
challenging samples are shown in Fig 5. It can be seen that these prostate images have fuzzy boundaries and the
pixel intensity distributions are inhomogeneous both inside and outside of the prostate. Besides, both prostate and
nonprostate regions have similar contrast and intensity distributions. All of these phenomenons make the segmentation difficult. \vspace{2mm}

As shown in the second column in Fig 5. FCN model failed to obtain satisfactory result, though the model could detect part of prostate. However,
for the details of prostate, for instance, the boundaries, the network can not assign the label to each pixel accurately. \vspace{0mm}

In U-Net model, the label has be assigned to each pixel and has improved the segment accurate as shown in the third column in Fig 5. However,
the network always make a mistake when assigning the label to nonprostate regions. Besides, for the boundaries information, the network cannot work well. At last, the
segmentation results lose some important information and there still exist some errors.\vspace{2mm}

The results of deeply supervised CNN are shown in the fourth column of Fig 5. The fuzzy boundaries are well detected by our proposed method. Besides, the segmentation
boundary are more continuous and smooth than the competing method. It can be proved that additional supervised layers are useful for the texture and boundaries of prostate.

\begin{table}[!t]
\centering
\caption{Quantitative comparison between the proposed approach with other methods}
\label{table_delay}
\begin{IEEEeqnarraybox}[\IEEEeqnarraystrutmode\IEEEeqnarraystrutsizeadd{2pt}{0pt}]{x/r/Vx/r/v/r/v/r/x}
\IEEEeqnarraydblrulerowcut
\\
&&&&\IEEEeqnarraymulticol{5}{t}{DSC}&
\\
&\hfill\raisebox{-2pt}[0pt][0pt]{method}
\hfill&&\IEEEeqnarraymulticol{7}{h}{}%
\IEEEeqnarraystrutsize{0pt}{0pt}
\\
&&&&\hfill{meanDSC}\hfill&&\hfill{medianDSC}\hfill&&\hfill{maximumDSC}\hfill&
\IEEEeqnarraystrutsizeadd{0pt}{0pt}\\
\IEEEeqnarraydblrulerowcut\\
&{Our-model}&&&0.885&&0.945&&0.985&\\
&{U-Net}&&&0.865&&0.940&&0.969&\\
&{FCN}&&&0.759&&0.832&& 0.918&\\
\IEEEeqnarraydblrulerowcut\\
\end{IEEEeqnarraybox}
\end{table}

\subsubsection{Quantitative Comparison}
The statistical results of the three methods are shown in table I. We use three ways to evaluate the results.
From table I we can see, the average of DSC values of our model , U-Net and FCN are 0.885 ,0.865 and 0.759  respectively.
Besides,all of the average, median and maximum DSC values of our method are the highest.  So it can also be proved that the proposed method obtains significant
improvement on the prostate segmentation compare with the other methods. These improvement can be attributed to the proposed network more deeper and the additional
supervised layers. During training, the additional supervised layers  make a strong constrain on the network. And these supervised layers can solve the problem that the fuzzy boundaries and the pixel intensity distributions are inhomogeneous both inside and outside of the prostate.\vspace{2mm}
\begin{figure*}
\centering

\includegraphics[width=15cm,height=10cm]{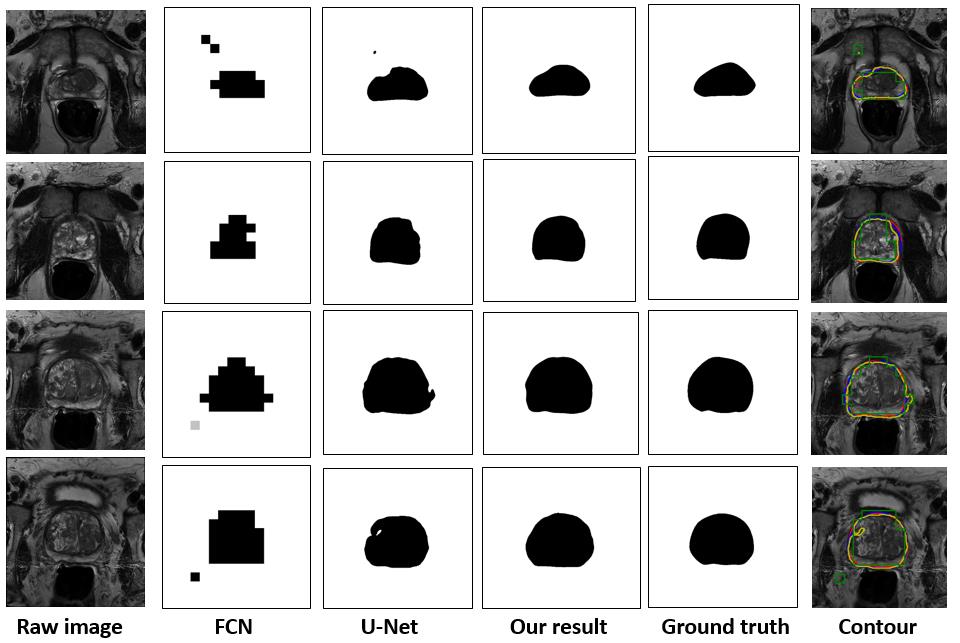}
\caption{Segmentation results. From left to right are Raw image,the segmentation results of FCN,the segmentation results of U-Net,the segmentation results of Deeply-supervised CNN,
Ground Truth,Segmentation results respectively. The blue, red ,green and yellow contour denote ground truth, our result, the result of FCN  and the result of U-Net respectively. }
\end{figure*}

\begin{figure*}
\centering

\includegraphics[width=15cm,height=7cm]{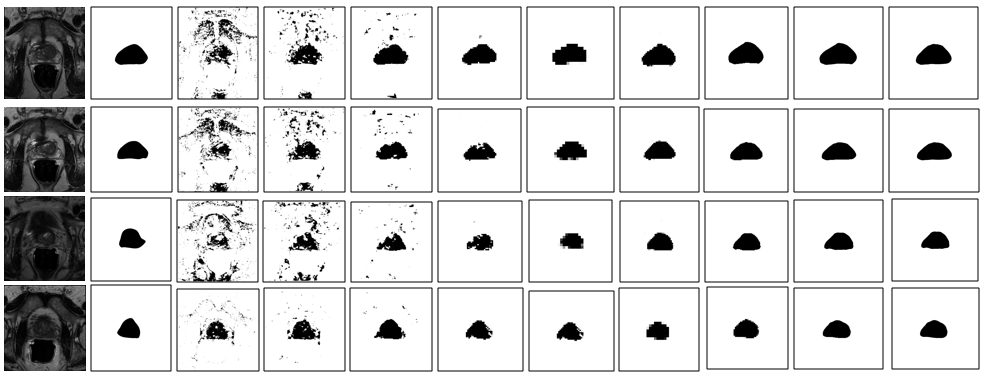}
\caption{Different deep supervised layers' segmentation results. From left to right are Raw image,Ground Truth,the segmentation results of deep supervised layers respectively. }
\end{figure*}
\subsubsection{Discussion}
Our results seem to yield a solid evidence that imposing a deeply supervised method during training the network is a viable
method for improving neural network's performances for medical images segmentation. During training, all of these supervised layers will supervise the process
of training and reduce the lose of prostate information. And due to the depth of the network's large scale, the additional
supervised layers can provide gradients information for early stage which resolves the problem of gradients diffusion. As shown
in Fig 6 . It can be seen that the different supervised layers can detect different textures. The later layers are
 closer to ground truth and the early supervised layers possess more information. So the supervised layers located in
 early stages could provide the information which was lost during training for the later stage.

\section{Conclusion}
In this paper, we propose a deeply-supervised CNN that utilized the residual information to accurately segment the prostate MRI.
The proposed network is deeper and the number of parameters do not increase simultaneously compared with the traditional U-Net by applying 1x1
convolution. And the additional deep supervision plays a supervisory role during training the network. These supervised layers can avoid pixels' information loss to
some extent during training. For the prostate image, the amount of background and foreground pixels are quite unbalanced. As a result, the network which ignores
the segmentation parts and the output of network is strongly biased towards background. This result that the learning process get trapped in local minima and we can not obtain accurate results
finally. To resolve this problem, we apply the dice coefficient as the objective function. The results show that the proposed network improves the performance
of segmentation.

\section{Acknowledgement}
This work was supported in part by the Open Foundation of Basic Scientific Research Operating Expenses of Central-Level Public Academies and Institutes (CKWV2016380/KY), the National Natural Science Foundation of China under Grants 61471274£¬ the Natural Science Foundation of Hubei Province under Grants 2014CFB193 and  the Fundamental Research Funds for the Central Universities under Grants  2042016kf0152.





\bibliographystyle{IEEEtran}
%

\bibliographystyle{IEEEtran}

\begin{thebibliography}{1}
\bibitem{LeaningDe}
H.~Noh, S.~Hong, and B.~Han, ``Learning deconvolution network for semantic
  segmentation,'' \emph{Proceedings of the IEEE International Conference on
  Computer Vision},  vol. abs/1505.04366, 2015.

\bibitem{Baxt1992Use}
W.~G. Baxt, ``Use of an artificial neural network for the diagnosis of
  myocardial infarction.'' \emph{Annals of Internal Medicine},  vol. 115,
  no.~11, pp. 843--8, 1992.

\bibitem{Szegedy2014Going}
C.~Szegedy, W.~Liu, Y.~Jia, and P.~Sermanet,  ``Going deeper with
  convolutions,'' pp. 1--9, 2014.

\bibitem{Krizhevsky2012ImageNet}
A.~Krizhevsky, I.~Sutskever, and G.~E. Hinton, ``Imagenet classification with
  deep convolutional neural networks,'' \emph{Advances in Neural Information
  Processing Systems},  vol.~25, no.~2, p. 2012, 2012.

\bibitem{Simonyan2014Deep}
K.~Simonyan, A.~Vedaldi, and A.~Zisserman, ``Deep inside convolutional
  networks: Visualising image classification models and saliency maps,'' ''
  \emph{Computer Science}, 2014.

\bibitem{Dan2012Multi}
C.~Dan, U.~Meier, and J.~Schmidhuber, ``Multi-column deep neural networks for
  image classification,'' \emph{Proceedings / CVPR, IEEE Computer Society
  Conference on Computer Vision and Pattern Recognition. IEEE Computer Society
  Conference on Computer Vision and Pattern Recognition},  vol. 157, no.~10,
  pp. 3642--3649, 2012.

\bibitem{FULL}
S.~E. a. D.~T. Long, J., ``Fully convolutional networks for semantic
  segmentation.''  \emph{In Proceedings of the IEEE Conference on Computer
  Vision and Pattern Recognition}, pp. 3431--3440, 2015.

\bibitem{Zhang2015Deep}
W.~Zhang, R.~Li, H.~Deng, L.~Wang, W.~Lin, S.~Ji, and D.~Shen, ``Deep
  convolutional neural networks for multi-modality isointense infant brain
  image segmentation.'' \emph{Neuroimage},  vol. 108, pp. 214--224, 2015.

\bibitem{Simonyan2014Very}
K.~Simonyan and A.~Zisserman, ``Very deep convolutional networks for
  large-scale image recognition,'' '' \emph{Computer Science}, 2014.

\bibitem{He2014Spatial}
K.~He, X.~Zhang, S.~Ren, and J.~Sun, ``Spatial pyramid pooling in deep
  convolutional networks for visual recognition.'' \emph{IEEE Transactions on
  Pattern Analysis and Machine Intelligence},  vol.~37, no.~9, pp. 1904--16,
  2014.

\bibitem{Yan2011Adaptively}
P.~Yan, S.~Xu, B.~Turkbey, and J.~Kruecker, ``Adaptively learning local shape
  statistics for prostate segmentation in ultrasound.'' \emph{IEEE Transactions
  on Biomedical Engineering},  vol.~58, no.~3, pp. 633--641, 2011.

\bibitem{Dan2012Deep}
C.~C. Dan, A.~Giusti, L.~M. Gambardella, and Schmidhuber, ``Deep neural
  networks segment neuronal membranes in electron microscopy images,''
  \emph{Advances in Neural Information Processing Systems},  vol.~25, pp.
  2852--2860, 2012.

\bibitem{Kleesiek2016Deep}
J.~Kleesiek, G.~Urban, A.~Hubert, D.~Schwarz, K.~Maierhein, M.~Bendszus, and
  A.~Biller, ``Deep mri brain extraction: A 3d convolutional neural network for
  skull stripping.'' \emph{Neuroimage},  vol. 129, pp. 460--469, 2016.

\bibitem{Suk2014Hierarchical}
H.~I. Suk, S.~W. Lee, and D.~Shen, ``Hierarchical feature representation and
  multimodal fusion with deep learning for ad/mci diagnosis.''
  \emph{Neuroimage},  vol. 101, pp. 569--582, 2014.

\bibitem{Wu2016Scalable}
G.~Wu, M.~Kim, Q.~Wang, B.~C. Munsell, and D.~Shen, ``Scalable high-performance
  image registration framework by unsupervised deep feature representations
  learning.'' \emph{IEEE Transactions on Biomedical Engineering},  vol.~63,
  no.~7, pp. 1--1, 2016.

\bibitem{Cire2013Mitosis}
D.~C. Cirean, A.~Giusti, L.~M. Gambardella, and J.~Schmidhuber, ``Mitosis
  detection in breast cancer histology images with deep neural networks.'' in
  \emph{Medical Image Computing and Computer-assisted Intervention: Miccai
  International Conference on Medical Image Computing and Computer-assisted
  Intervention}, 2013, pp. 411--8.

\bibitem{Drozdzal2016The}
M.~Drozdzal, E.~Vorontsov, G.~Chartrand, S.~Kadoury, and C.~Pal, \emph{The
  Importance of Skip Connections in Biomedical Image Segmentation}.\hskip 1em
  plus 0.5em minus 0.4em\relax Springer International Publishing, 2016.

\bibitem{Lefei}
L.~Zhang, Q.~Zhang, L.~Zhang, D.~Tao, X.~Huang, and B.~Du, ``Ensemble manifold
  regularized sparse low-rank approximation for multiview feature embedding,''
  \emph{Pattern Recognit.},  vol.~48, no.~10, pp. 3102--3112, 2015.

\bibitem{Shen2001HAMMER}
D.~Shen and C.~Davatzikos, ``Hammer: hierarchical attribute matching mechanism
  for elastic registration.'' \emph{IEEE Transactions on Medical Imaging},
  vol.~21, no.~11, pp. 29--36, 2001.

\bibitem{Liao2013Representation}
S.~Liao, Y.~Gao, A.~Oto, and D.~Shen, ``Representation learning: a unified deep
  learning framework for automatic prostate mr segmentation.'' in \emph{Medical
  Image Computing and Computer-assisted Intervention: Miccai International
  Conference on Medical Image Computing and Computer-assisted Intervention},
  2013, pp. 254--61.

\bibitem{Yan2010Discrete}
P.~Yan, S.~Xu, B.~Turkbey, and J.~Kruecker, ``Discrete deformable model guided
  by partial active shape model for trus image segmentation.'' \emph{IEEE
  transactions on bio-medical engineering},  vol.~57, no.~5, pp. 1158--66,
  2010.

\bibitem{Toth2012Multifeature}
R.~Toth and A.~Madabhushi, ``Multifeature landmark-free active appearance
  models: Application to prostate mri segmentation,'' \emph{IEEE Transactions
  on Medical Imaging},  vol.~31, no.~8, pp. 1638--50, 2012.

\bibitem{Xie2015Holistically}
S.~Xie and Z.~Tu, ``Holistically-nested edge detection,'' in \emph{IEEE
  International Conference on Computer Vision}, 2015, pp. 1395--1403.

\bibitem{Cheng}
L.~L. Cheng~R, Roth H~R, ``Active appearance model and deep learning for more
  accurate prostate segmentation on mri.'' \emph{International Society for
  Optics and Photonics},  vol. 100, pp. 978\,421--978\,421--9, 2016.

\bibitem{Chen2016DCAN}
H.~Chen, X.~Qi, L.~Yu, and P.~A. Heng, ``Dcan: Deep contour-aware networks for
  accurate gland segmentation,'' '' \emph{arXiv:1604.02677}, 2016.

\bibitem{Long2015Fully}
J.~Long, E.~Shelhamer, and T.~Darrell, ``Fully convolutional networks for
  semantic segmentation,'' in \emph{IEEE Conference on Computer Vision and
  Pattern Recognition}, 2015, pp. 1337--1342.

\bibitem{Lee2014Deeply}
C.~Y. Lee, S.~Xie, P.~Gallagher, Z.~Zhang, and Z.~Tu, ``Deeply-supervised
  nets,'' '' \emph{Eprint Arxiv}, pp. 562--570, 2014.

\bibitem{Ronneberger2015U}
O.~Ronneberger, P.~Fischer, and T.~Brox, ``U-net: Convolutional networks for
  biomedical image segmentation,'' '' \emph{International Conference on Medical
  Image Computing and Computer-Assisted Intervention.}

\bibitem{Gu2015Recent}
J.~Gu, Z.~Wang, J.~Kuen, L.~Ma, A.~Shahroudy, B.~Shuai, T.~Liu, X.~Wang, and
  G.~Wang, ``Recent advances in convolutional neural networks,'' ''
  \emph{Computer Science}, 2015.

\bibitem{Duan2010A}
C.~Duan, Z.~Liang, S.~Bao, and H.~Zhu, ``A coupled level set framework for
  bladder wall segmentation with application to mr cystography,'' \emph{IEEE
  Transactions on Medical Imaging},  vol.~29, no.~3, pp. 903--915, 2010.

\bibitem{2016arXiv160502688full}
P.~Charles, ``Project title,'' '' \emph{GitHub repository}, 2013.

\end{thebibliography}

\end{document}